\documentclass[sigconf]{acmart}
\AtBeginDocument{%
  }

\setcopyright{acmlicensed}
\copyrightyear{2025}
\acmYear{2025}
\acmDOI{XXXXXXX.XXXXXXX}
\acmConference[SIGSPATIAL ’25]{International Conference on Advances in Geographic Information Systems}{November 03 -- 05, 2025}{Minneapolis, MN, USA}

\acmISBN{978-1-4503-XXXX-X/2018/06}

\usepackage{import}

\begin{document}

\title{The Maximum Coverage Model and Recommendation System for UAV Vertiports Location Planning [Applications]}

\author{Chunliang HUA}
\authornote{This work was done when Chunliang HUA was interns at IDEA.}
\email{chunlianghua@seu.edu.cn}
\orcid{0009-0007-3082-2899}
\affiliation{%
  \institution{School of Information Science and Engineering, Southeast University}
  \city{Nanjing}
  \country{China}
}

\author{Xiao HU}
\email{huxiao1@idea.edu.cn}
\affiliation{%
  \institution{Lower AirSpace Economy Research Institute, International Digital Economy Academy}
  \city{Shenzhen}
  \country{China}}

\author{Jiayang SUN}
\email{sunjiayang@idea.edu.cn}
\affiliation{%
  \institution{Lower AirSpace Economy Research Institute, International Digital Economy Academy}
  \city{Shenzhen}
  \country{China}}

\author{Zeyuan YANG}
\email{yangzeyuan@idea.edu.cn}
\affiliation{%
  \institution{Lower AirSpace Economy Research Institute, International Digital Economy Academy}
  \city{Shenzhen}
  \country{China}}





\renewcommand{\shortauthors}{Chunliang et al.}

\begin{abstract}

As urban aerial mobility (UAM) infrastructure development accelerates globally, cities like Shenzhen are planning large-scale vertiport networks (e.g., 1,200+ facilities by 2026). Existing planning frameworks remain inadequate for this complexity due to historical limitations in data granularity and real-world applicability. This paper addresses these gaps by first proposing the Capacitated Dynamic Maximum Covering Location Problem (CDMCLP), a novel optimization framework that simultaneously models urban-scale spatial-temporal demand, heterogeneous user behaviors, and infrastructure capacity constraints. Building on this foundation, we introduce an Integrated Planning Recommendation System that combines CDMCLP with socio-economic factors and dynamic clustering initialization. This system leverages adaptive parameter tuning based on empirical user behavior to generate practical planning solutions. Validation in a Chinese center city demonstrates  the effectiveness of the new optimization framework and recommendation system. Under the evaluation and optimization of CDMCLP, the quantitative performance of traditional location methods are exposed and can be improved by 38\%--52\%, while the recommendation system shows user-friendliness and the effective integration of complex elements. By integrating mathematical rigor with practical implementation considerations, this hybrid approach bridges the gap between theoretical location modeling and real-world UAM infrastructure planning, offering municipalities a pragmatic tool for vertiport network design.
\end{abstract}

\ccsdesc[500]{Applied computing~Multi-criterion optimization and decision-making}
\ccsdesc[500]{Human-centered computing~Wireframes}

\keywords{UAV vertiport, Maximum coverage problem, Operation optimization, Recommendation system}


\maketitle

\section{Introduction}

The rapid growth of transportation demand coupled with increasingly complex traffic scenarios has created unprecedented challenges for ground transportation systems~\cite{yannis_transport_2022}. In urban core areas, traffic congestion and high management costs have become persistent obstacles to sustainable development, while rural regions, particularly hilly terrains, face infrastructure limitations despite abundant land resources, primarily due to economic viability concerns. This spatially asymmetric transportation dilemma has not only driven continuous innovations in technologies like autonomous driving and high-speed rail, but also catalyzed a revolutionary conceptual shift toward three-dimensional airspace utilization through Urban Air Mobility (UAM). Current technological trajectories demonstrate that UAM represents more than just a complementary solution to ground transportation, it possesses transformative potential for urban spatial paradigms. In logistics applications, drones have already demonstrated significant advantages over traditional human delivery systems, with Shenzhen's drone delivery pilot program showing over 60\% improvement in short-haul freight efficiency. For urban commuting, electric vertical takeoff and landing (eVTOL) vehicles offer three-dimensional routing solutions that effectively bypass ground congestion. According to Boston Consulting Group projections, this technology could reduce core urban commute times by 40\% within the next decade. More profoundly, this ``de-terrestrial'' transportation model may fundamentally reshape traditional urban management frameworks, much like how military drones have revolutionized battlefield operations~\cite{crunkhorn_chinas_2025,zhou_unmanned_2025}. The infrastructure-independent mobility characteristics of UAVs are creating entirely new urban spatial organization principles.

Central to this transformation is the UAV vertiport infrastructure network, which provides critical operational support including aircraft takeoff/landing, energy replenishment, and equipment storage. These facilities serve as foundational nodes enabling multi-industry low-altitude flight activities through shared resource allocation. Among vertiport development considerations, location planning emerges as a pivotal factor ensuring safe, efficient, and sustainable low-altitude economy growth. This requires comprehensive optimization across operational efficiency, economic viability, and environmental sustainability parameters. However, while substantial research has focused on vertiport architectural design~\cite{johnston_take_2020}, empirical studies on UAM vertiport location planning remain scarce and largely theoretical. This research gap primarily stems from UAM's nascent status and historical data deficiencies regarding demand patterns. Until recently, no global city had constructed a large-scale UAV vertiport network, creating a vacuum in practical implementation models. However, this situation is undergoing fundamental change. For example, the Shenzhen Municipal Government has established an ambitious target of constructing over 1,200 UAV vertiports by 2026~\cite{noauthor_shenzhen_2024}. This strategic initiative not only underscores urgent practical needs but also provides critical real-world foundations for academic research. The development of location models that balance operational efficiency, economic feasibility, and environmental compatibility has thus become a pivotal research challenge for low-altitude economy advancement.

To address these challenges, this work proposes a three-tiered framework integrating theoretical innovation with practical deployment.
Our solution unfolds in three stages. First, we introduce the capacitated dynamic maximum covering location problem (CDMCLP) to evaluate UAV vertiport network performance. By modeling demand spillover effects under capacity constraints, this approach quantifies both service coverage and operational efficiency for urban air mobility networks. Second, we develop a scalable optimization framework combining initialization heuristics with a tailored algorithm. Through strategic network initialization followed by model-driven optimization, our approach efficiently generates high-performance vertiport configurations. The methodology demonstrates robustness in large-scale scenarios, as shown in Shenzhen's case study requiring selection of 2,000 vertiports from 140,000 candidate locations, where our simplified yet effective operation optimization strategy maintains computational tractability. Third, we enhance practical applicability through a socio-technical integration approach. Our UAV vertiport recommendation system synthesizes three dimensions: (1) model-generated performance metrics, (2) socio-economic factors (population density, land values), and (3) existing infrastructure data. This hybrid framework provides planners with data-driven recommendations while preserving human decision-making authority. In summary, the contributions of this paper are as follows:
\begin{itemize}
\item Formulating large-scale vertiport planning as CDMCLP, enabling simultaneous assessment of service coverage and operational efficiency.
\item Developing a scalable optimization framework combining initialization heuristics with efficient operation optimization.
\item Creating a socio-technical recommendation system integrating technical performance metrics with socio-economic considerations.
\end{itemize}

The remainder of this paper is organized as follows: Section \ref{sec2} reviews the related morphology studies. Section \ref{sec3} provides the details of the methodology. The experimental results and analysis are presented in Section \ref{sec4}. The paper ends with a discussion in Section \ref{sec5} and conclusion in Section \ref{sec6}.

\section{Related work}
\label{sec2}
\subsection{UAV vertiports location planning}

Current research on UAV vertiport planning primarily falls into two categories: demand-based clustering approaches and mathematical programming methods.
Demand-Based Clustering Approaches dominate early UAM vertiport planning studies. Pioneering works applied K-means clustering to optimize vertiport locations in Seoul \cite{lim_selection_2019}, North California \cite{tarafdar_urban_2019}, and South Florida \cite{wei_optimal_2020}, focusing on minimizing travel distances between users and vertiports. More sophisticated techniques emerged subsequently: Rajendran et al. \cite{rajendran_insights_2019} introduced a multimodal transportation-based warm start (MTWS) strategy for clustering initialization, while Amitanand and Sinha \cite{amitanand_sinha_study_2023} employed CLARA algorithm in New York City case studies. These methods excel in computational efficiency but inherently assume static demand patterns without capacity constraints. 
Mathematical Programming Approaches address these limitations through formal optimization models. Wang et al. \cite{kai_vertiport_2022} developed a multi-level framework integrating strategic deployment, tactical operations, and demand dynamics with global convergence guarantees. Willey et al. \cite{willey_method_2021} incorporated eVTOL technical parameters to minimize average travel time, while Yu et al. \cite{jiang_vertiport_2025} optimized network design through three-dimensional criteria considering on-demand mobility (ODM) and regular shuttle (RS) demands. Most recently, Yu et al. \cite{yu_vertiport_2023} established an air-ground traffic integration model to reduce urban congestion. However, these complex formulations often face scalability challenges in large-scale implementations.

Despite the methodological diversity in UAV vertiport planning, existing studies face three critical limitations. First, the scale-reality gap arises as most works optimize networks of only dozens of vertiports (e.g., \cite{lim_selection_2019,tarafdar_urban_2019}), which starkly contrasts with the practical demands of megacities like Shenzhen aiming for 2,000 vertiports. Second, demand misalignment persists due to an overemphasis on medium-distance commercial eVTOL applications, diverging from current UAM development trajectories that prioritize short-range, high-frequency operations. Third, data deficiency plagues most models, as the absence of real-world demand patterns forces reliance on simplistic or unrealistic assumptions. While deep learning has shown promise in related domains, such as Raczcyki et al.'s \cite{raczycki_transfer_2021} transfer-learning approach for bicycle-sharing station planning using OpenStreetMap data, or Wang et al.'s \cite{wang_leveraging_2020} CNN-based ESLE model for geospatial embedding, these methods remain underexplored for vertiport planning due to UAM data scarcity. These gaps collectively highlight the urgent need for a framework addressing scalable optimization under capacity constraints, dynamic demand modeling, and practical deployment considerations in large-scale urban environments—all core objectives of our proposed methodology.

\subsection{Maximum covering location problem (MCLP)}

The Maximum Covering Location Problem (MCLP) has emerged as a foundational framework for modeling urban air mobility (UAM) infrastructure due to its intuitive formulation, computational tractability, and ability to capture complex interactions between facilities and dynamic demand patterns. Originally proposed by Church and ReVelle \cite{church_maximal_1974} to maximize customer coverage with fixed facility numbers, MCLP has become a cornerstone of location optimization despite its NP-hard complexity \cite{hartmanis_computers_1982}. Classical heuristic approaches include greedy heuristics \cite{downs_exact_1996}, Lagrangian relaxation \cite{galvao_lagrangean_1996}, simulated annealing \cite{murray_applying_1996}, and tabu search \cite{adenso-diaz_simple_1997}, while metaheuristics like genetic algorithms \cite{jaramillo_use_2002,fazel_zarandi_large_2011} and, more recently, graph convolutional networks \cite{zhang_dual_2024} demonstrate ongoing methodological evolution.

The MCLP's adaptability has spawned numerous variants addressing domain-specific constraints: budget limits \cite{khuller_budgeted_1999}, generalized coverage \cite{cohen_generalized_2008}, collinear facility requirements \cite{bhattacharya_new_2013}, partial coverage \cite{peker_p-hub_2015}, and fuzzy capacitated formulations \cite{atta_solving_2022}. Building on these advancements, we introduce the capacitated dynamic MCLP (CDMCLP) to address UAM's unique challenges. Our model extends classical MCLP by incorporating time-varying demand patterns, capacity constraints at vertiports, and spatiotemporal coverage dynamics—features absent in prior variants yet critical for realistic urban deployment. This adaptation bridges a key gap in existing literature, enabling scalable optimization of high-density vertiport networks while maintaining computational feasibility.

\section{Methodology}
\label{sec3}

\subsection{Problem Formulation}
\begin{figure}[htbp]
  \centering
  \includegraphics[width=\linewidth]{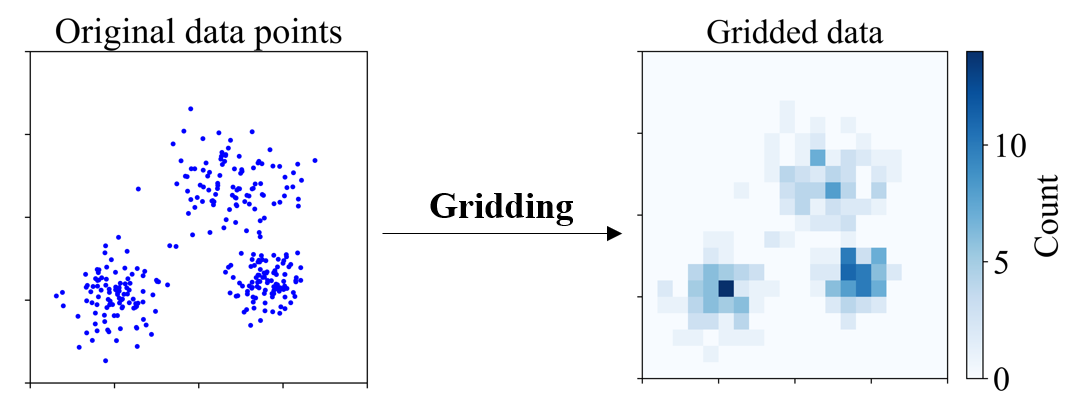}
  \caption{Example of gridding: the left panel shows Cartesian coordinate system with scattered points, while the right panel displays corresponding heatmap.}
  \Description{}
  \label{fig3-1}
\end{figure}

The proposed CDMCLP operates on a discretized spatiotemporal grid framework (Figure~\ref{fig3-1}), enabling efficient matrix-based computations via libraries such as PyTorch and NumPy. Time and space are represented by integer indices: $ t \in \{0, 1, \ldots, T-1\} $ for temporal resolution, and $ i, j \in \{0, 1, \ldots, M-1\} \times \{0, 1, \ldots, N-1\} $ for spatial resolution. Regions are denoted as $ A_{i,j} $, forming a grid of $ M \times N $ spatial units.

\begin{figure*}[!htb]  
    \centering  
    \includegraphics[width=\textwidth]{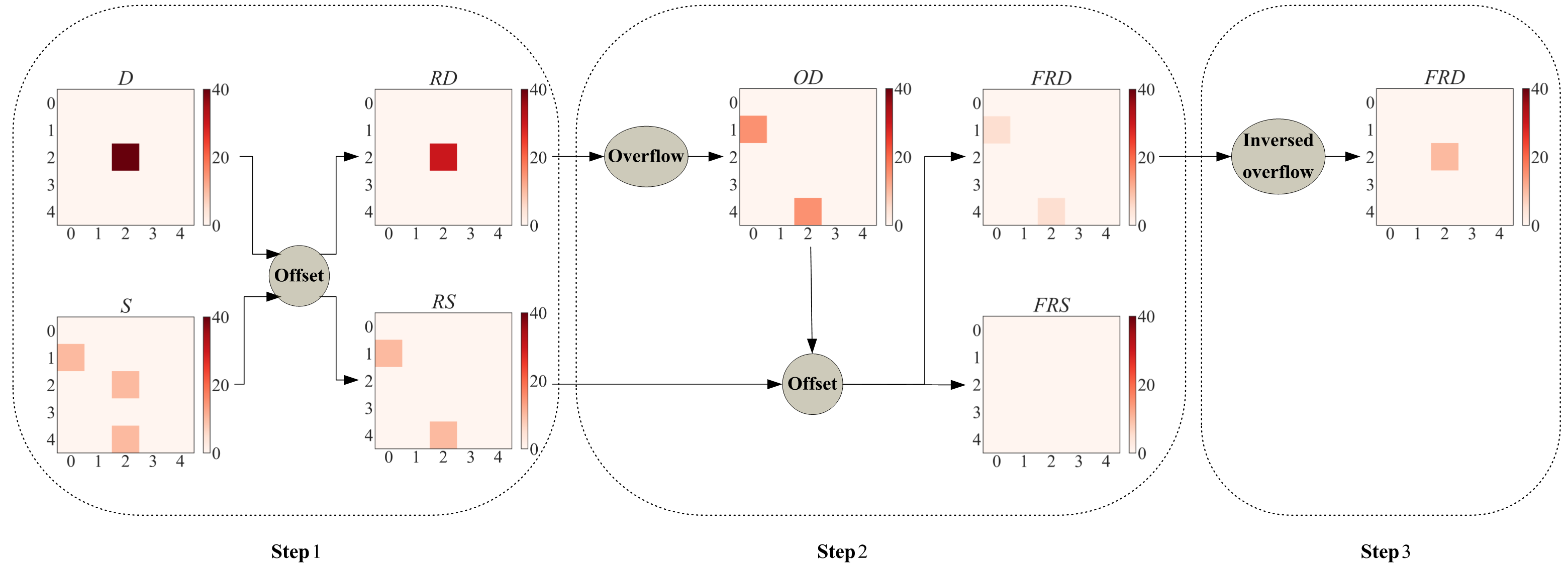}  
    \caption{Framework of algorithm $ G $: Heatmaps illustrate sequential transformations in supply-demand matching.} 
  \Description{}
    \label{fig3-2}  
\end{figure*}

\noindent\textbf{Demand:} The demand points in our framework possess both spatial and temporal attributes. To capture this dynamic, we represent demand distribution as a three-dimensional tensor $ D \in \mathbb{Z}_{\geq 0}^{T \times M \times N} $, where $ D_{t,i,j} $ quantifies the number of demand points in region $ A_{i,j} $ at time interval $ t $. This formulation explicitly models temporal variability in UAM demand patterns.

\noindent\textbf{Supply:} Supply capacity, in contrast, is treated as a static resource tied to facility locations. We define the supply capacity matrix $ S \in \mathbb{Z}_{\geq 0}^{M \times N} $, with $ S_{i,j} $ representing the total service capacity in region $ A_{i,j} $ as follows: 
\begin{equation}
    S_{i,j} = V_{i,j} * p \ ,
    \label{eq1}
\end{equation}
where \( V_{i,j} \) represents the number of facilities in region \( A_{i,j} \) and \( p \) is the supply capacity that a site can provide per unit time.

\noindent\textbf{Feasibility Constraints:} To ensure physical realizability of the facility layout, $ S $ must satisfy three constraints:  
\begin{enumerate}
    \item Total Supply Constraint:
    \begin{equation}
    \sum_{i=0}^{M-1} \sum_{j=0}^{N-1} S_{i,j} = p * c \ ,
    \label{eq2}
    \end{equation}
    where $c$ is the total number limit of facilities. This constraint indicates the total upper limit of supply capacity.
    \item Capacity Granularity:
    \begin{equation}
    \forall i \in \{0, 1, \ldots, M - 1\}, \forall j \in \{0, 1, \ldots, N - 1\}, S_{i,j} \bmod p = 0 \ ,
    \label{eq3}
    \end{equation}
    which ensures that the supply capacity in any region can only be an integer multiple of $p$.
    \item Non-Negativity:
    \begin{equation}
    \forall i \in \{0, 1, \ldots, M - 1\}, \forall j \in \{0, 1, \ldots, N - 1\}, S_{i,j} \geq 0 \ ,
    \label{eq4}
    \end{equation}
    which indicates that the supply capacity value in any region cannot be negative.
\end{enumerate}

To model the complex interaction between supply and demand, we introduce a dynamic matching algorithm G (as shown in Figure~\ref{fig3-2}). Unlike traditional binary coverage models, which assume static and unrestricted service availability, G explicitly simulates user behavior in seeking services under realistic capacity constraints through three sequential stages.

The first stage is on local clearance, where demand and supply within the same region $ A_{i,j} $ are matched. This ensures that users prioritize utilizing nearby resources before seeking alternatives. Residual demand ($ RD_t $) and residual supply ($ RS_t $) are calculated as:  
   $$
   RD_t = \max(D_t - S, 0),\quad RS_t = \max(S - D_t, 0).
   $$  
In the second stage, demand redistribution is performed. The residual demand $ RD_t $ ``overflows'' to neighboring regions within the UAV vertiport service radius r (a hyper-parameter reflecting operational range limits). These overflowed demands seek service in regions that still retain residual supply $RS_t$. 
This generates overflowed demand ($ OD_t $), which is then matched with $ RS_t $:  
   $$
   FRD_t = \max(OD_t - RS, 0),\quad FRS_t = \max(RS_t - OD_t, 0).
   $$  
Finally, the third stage performs backward mapping to trace the final residual demand $ FRD_t $ back to its origin regions. While this step does not alter the numerical values of $ FRD_t $, it preserves the spatial context of unmet demand, ensuring that subsequent optimization processes (e.g., facility repositioning) retain geographic fidelity.

In summary, the algorithm’s input consists of the demand tensor $D$ and supply matrix $S$, with outputs $ FRD_t $ and $ FRS_t $ capturing unmet demand and unused capacity at each time step. The relationship is formally expressed as:
\begin{equation}
    (FRD_t, FRS_t) = G(D_t, S)
    \label{eq5}
\end{equation}
By explicitly modeling spatiotemporal dynamics, capacity limits, and user behavior, this formulation enables scalable optimization of large-scale UAM networks while maintaining practical relevance to real-world deployment scenarios.

\subsection{Optimization}
Our optimization objective is defined as minimizing the total unmet demand over all time intervals and spatial regions:
\begin{equation}
    \min \sum_{t=0}^{T-1} \sum_{i=0}^{M-1} \sum_{j=0}^{N-1} FRD_{t,i,j}
    \label{eq6}
\end{equation}
The algorithm operates through three core mechanisms leveraging $FRD$ and $FRS$:
\begin{itemize}
    \item Demand-Driven Supply Addition: $FRD$ is summed across the temporal dimension to aggregate unmet demand over time. A convolution kernel filled with ones is then applied to spatially smooth the demand distribution. The region exhibiting the highest smoothed value is selected to receive an additional supply capacity p. Intuitively, this corresponds to establishing a new UAV vertiport in areas experiencing the most severe and sustained demand-supply imbalance.
    \item Supply Reduction in Underutilized Regions: $FRS$ is similarly summed over time to identify regions with chronically underutilized capacity. The location with the highest residual supply value has its capacity reduced by $p$, effectively decommissioning a UAV vertiport in zones where service availability consistently exceeds demand. This step prevents resource wastage and reallocates capacity to more critical areas.
    \item Tabu List for Stability:
    To avoid oscillatory adjustments and ensure convergence, a tabu list mechanism is implemented. Any region that undergoes back-to-back modifications (e.g., a UAV vertiport removed followed by a new addition, or vice versa) is added to the tabu list, temporarily prohibiting further optimization actions on its capacity. This heuristic restriction mimics tabu search strategies, preventing redundant or destabilizing updates while maintaining computational simplicity.
\end{itemize}
This iterative process combines greedy heuristics (prioritizing high-demand regions) with tabu-like memory structures, though its implementation remains deliberately lightweight. Despite its simplicity, the algorithm rigorously satisfies all constraints ~\eqref{eq2}–\eqref{eq4} and demonstrates empirical effectiveness in large-scale experiments. Its computational efficiency, requiring minimal matrix operations, makes it a practical baseline for real-world deployments, particularly in scenarios demanding rapid re-optimization due to dynamic demand patterns or operational constraints.

\subsection{UAV vertiport location recommendation system}
\label{sec3-3}
\begin{figure*}[!htb]
\centering
\includegraphics[width=0.9\textwidth]{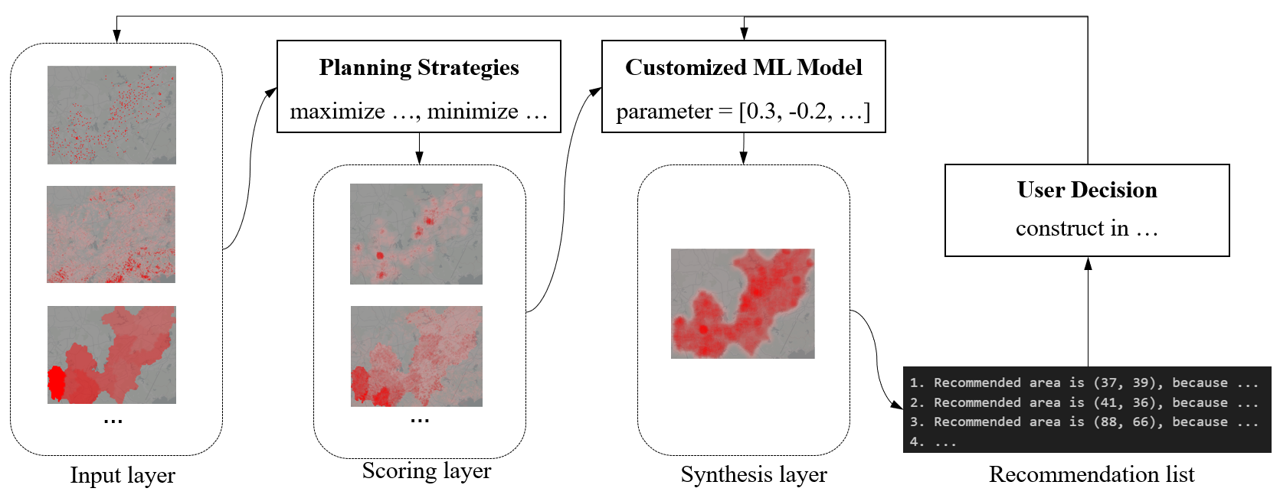}
\caption{Recommendation system framework: Input layer (left), scoring layer (middle), synthesis layer (right), and feedback loop.}
  \Description{}
\label{fig3-3}
\end{figure*}

\begin{figure}[h]
\centering
\includegraphics[width=\linewidth]{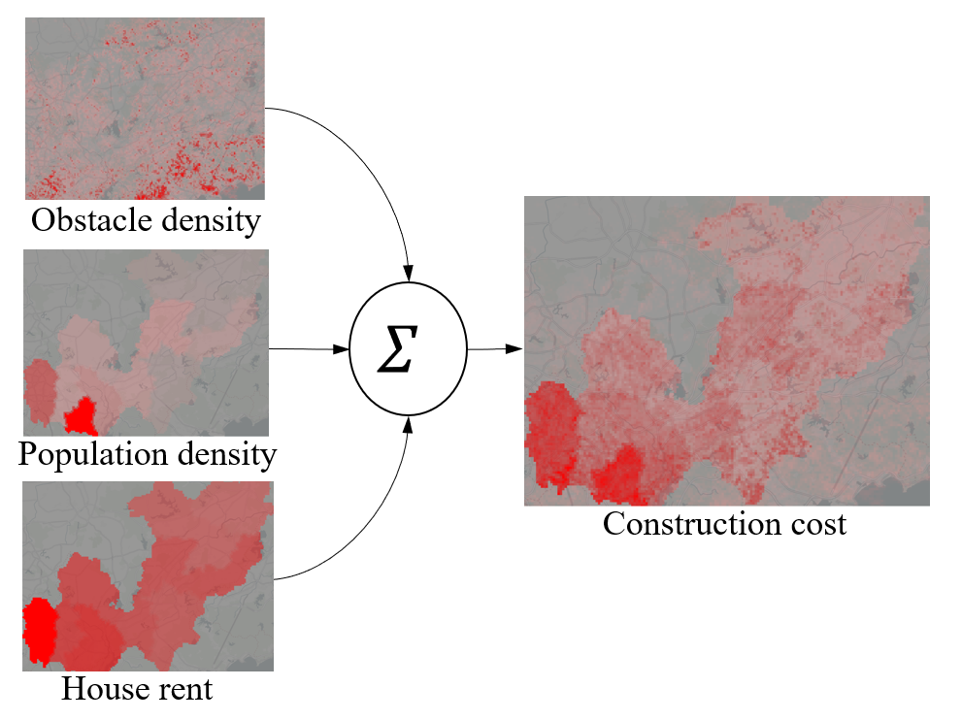}
\caption{Construction cost calculation pipeline: Normalized inputs (left) are summed and rescaled to generate a composite score map (right).}
  \Description{}
\label{fig3-4}
\end{figure}

The UAV Vertiport Location Recommendation System determines the UAV vertiport location once it obtains the optimization result. This system comprises three hierarchical layers and a feedback mechanism, as depicted in Figure \ref{fig3-3}:
\begin{itemize}
\item Input Layer: This layer aggregates heterogeneous data sources, such as real-time demand patterns, geographic constraints, and socio-economic factors, forming the foundation for subsequent analytical stages.
\item Scoring Layer: The input data is processed through parallel computation graphs in this layer, generating multiple $M\times N$ scoring matrices. Each matrix encodes region-specific scores, indicating the suitability for UAV vertiport placement under distinct planning strategies.
\item Synthesis Layer: All scoring matrices are integrated here using a custom machine learning (ML) model inspired by logistic regression. The model performs weighted summation followed by a sigmoid activation to produce a comprehensive scoring matrix. While its forward pass resembles logistic regression, its parameters are user-initialized and iteratively refined based on interaction logs.
\end{itemize}
The system recommends diverse, high-scoring regions to users, who retain final decision-making authority. User selections are fed back into the input layer and used to update synthesis-layer parameters, enabling adaptive learning over time.
Currently, the system incorporates four complementary strategies:
\begin{itemize}
\item \textbf{Demand Satisfaction Maximization:} This strategy uses the proposed CDMCLP to sum $FRD$ across time and spatially smooth it via a convolution kernel, prioritizing regions with persistent undersupply. To minimize latency during user interactions, the input demand matrix is preprocessed into a static $1 \times M \times N$ matrix by gridding a high-performance UAV vertiport network, akin to knowledge distillation.
\item \textbf{Coverage Area Maximization:} This approach scores regions based on the number of uncovered neighboring zones, with higher scores corresponding to locations that maximize service reach by filling gaps in the existing vertiport network.
\item \textbf{Air-Ground Connectivity Time Minimization:} This strategy quantifies the reduction in travel time between UAV vertiports (set A) and subway stations (set B) using the formula:
\begin{equation}
\sum_{a \in A} \min_{b \in B} T_{a,b} \ ,
\label{eq7}
\end{equation}
where $T_{a,b}$ is the direct travel time from vertiport $a$ to station $b$. Scores reflect the marginal connectivity improvement from adding a vertiport to each region, incentivizing proximity to transit hubs.
\item \textbf{Construction Cost Minimization:} This method combines normalized metrics of obstacle density, population density, and housing rental prices (Figure \ref{fig3-4}). These factors are linearly aggregated and rescaled to produce a composite cost score, prioritizing regions with lower development barriers.
\end{itemize}
This multi-strategy framework balances competing objectives, including service efficiency, accessibility, and cost-effectiveness, while preserving human oversight. It ensures practical relevance to urban planning workflows.

\section{Experiment and Results}
\label{sec4}
\subsection{Study area}
\begin{figure}[h]
  \centering
  \includegraphics[width=0.98  \linewidth]{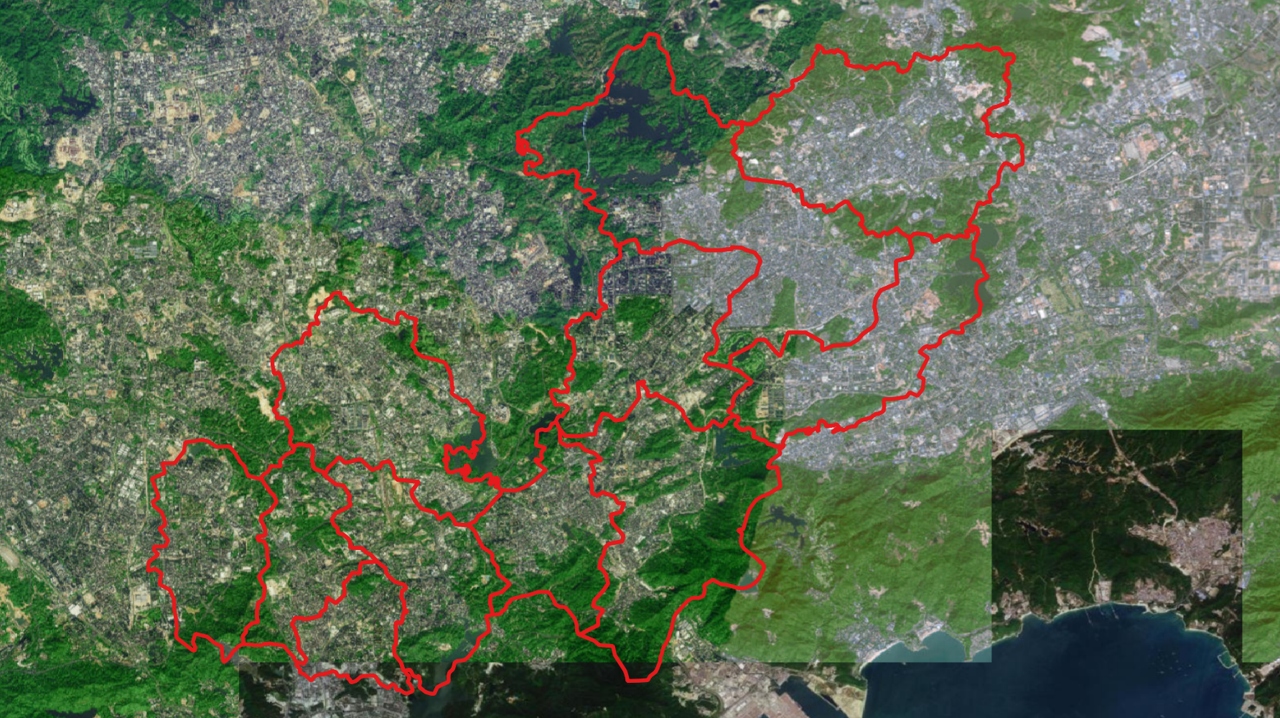}
  \caption{Satellite imagery of study area with a red border.}
  \Description{Satellite imagery of study area with a red border.}
  \label{fig4-1}
\end{figure}

A district of a Chinese center city, spanning approximately 388 square kilometers and home to around 4.1 million residents, serves as the primary study area for this research. As illustrated in the satellite map in Figure~\ref{fig4-1}, the district exhibits pronounced spatial heterogeneity, characterized by stark contrasts in population density and land-use patterns. The southwestern built-up areas, directly connected to bustling central districts, host population densities of up to 30,000 people per square kilometer, reflecting intense urban activity and demand for advanced mobility infrastructure. In contrast, the northeastern regions, which adjoin smaller cities, demonstrate significantly lower densities of approximately 2,000 people per square kilometer, indicative of suburban sprawl and less frequent transportation needs. Additionally, the northern lakes and southern mountainous zones--renowned scenic spots--experience seasonal demand surges due to tourism and limited ground infrastructure accessibility. This combination of hyper-dense urban cores, transitional suburban zones, and natural barriers makes the district a representative case study for evaluating large-scale UAV vertiport networks under complex real-world conditions.  

\begin{figure}[!htbp]
  \centering
  \includegraphics[width=0.98 \linewidth]{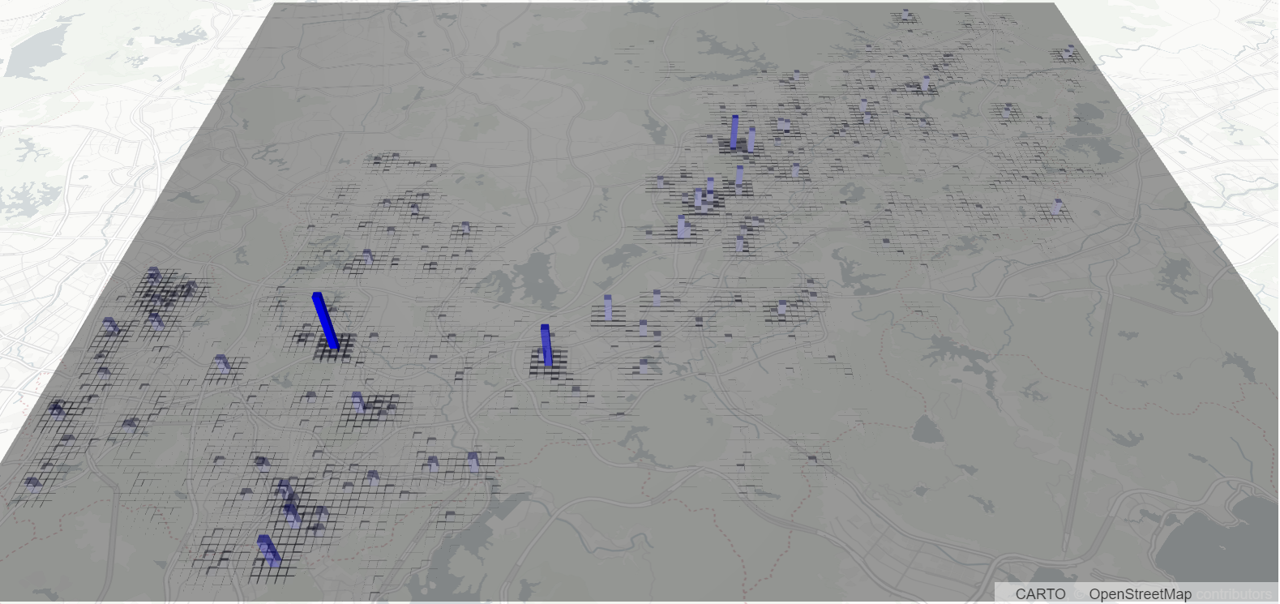}
  \caption{Spatiotemporal Demand Distribution Gridded into $ D \in \mathbb{Z}^{T \times M \times N} $: 3D Visualization of UAV Flight Origin-Destination Pairs in study area.}
  \Description{}
  \label{fig4-2}
\end{figure}

As part of Chinese advanced UAM pilot program, The study area provides abundant and accessible UAV flight trajectory data and urban contextual information. We acquired UAV flight records for November 2024--a period coinciding with peak tourism season--and extracted origin-destination (OD) pairs from flight start and end points to represent spatiotemporally resolved mobility demand. These OD pairs were gridded into a $ T \times M \times N $ demand tensor $ D $, as described in Section~\ref{sec3}, with the spatially aggregated result visualized in Figure~\ref{fig4-2}. This dataset captures both daily commuting patterns and recreational demand surges near scenic areas, providing a realistic foundation for evaluating vertiport network performance.

\begin{figure}[h]
\centering
\includegraphics[width=0.98\linewidth]{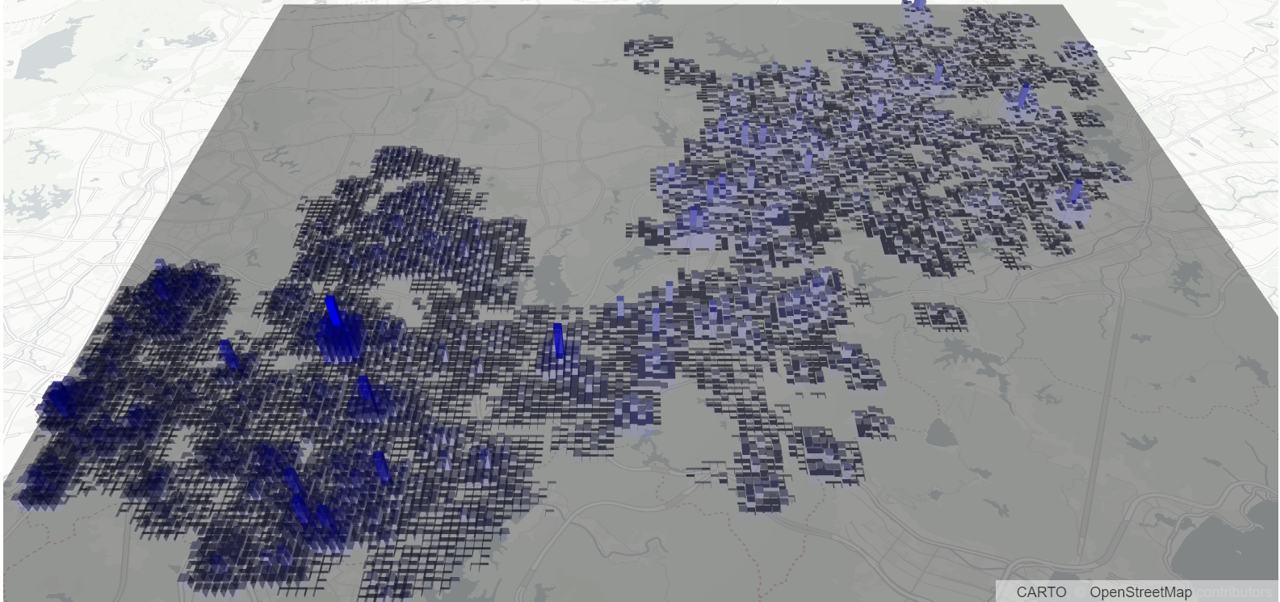}
\caption{Peak Demand Intensity Distribution: Maximum UAV Usage Frequency per Unit Time Across the study area}
\Description{A 3D map displaying gridded regions with color-coded and height-scaled blocks representing peak demand intensity (maximum flight count per hour) in the study area.}
\label{fig4-3}
\end{figure}

It is evident that the spatial distribution of UAV demand exhibits significant heterogeneity (Figure~\ref{fig4-2}), a pattern consistent with the nascent stage of low-altitude economy development in Chinese center city. To refine planning accuracy, this study explicitly incorporates temporal dynamics into demand analysis. Instead of aggregating demand across the entire observation period ($ T $), we extract the **peak demand intensity**--defined as the maximum number of origin-destination pairs recorded in any single time interval--for each spatial unit ($ A_{i,j} $). The resulting visualization (Figure~\ref{fig4-3}) reveals pronounced deviations from the static spatial distribution in Figure~\ref{fig4-2}, particularly in regions with transient demand spikes near tourist attractions and commuter hubs. This contrast underscores the criticality of spatiotemporal resolution: optimizing vertiport networks based solely on time-averaged demand risks misallocating infrastructure in areas where capacity requirements are highly time-dependent.

\subsection{Optimization results based on CDMCLP}

\begin{figure}[h]  
  \centering  
  \includegraphics[width=0.95\linewidth]{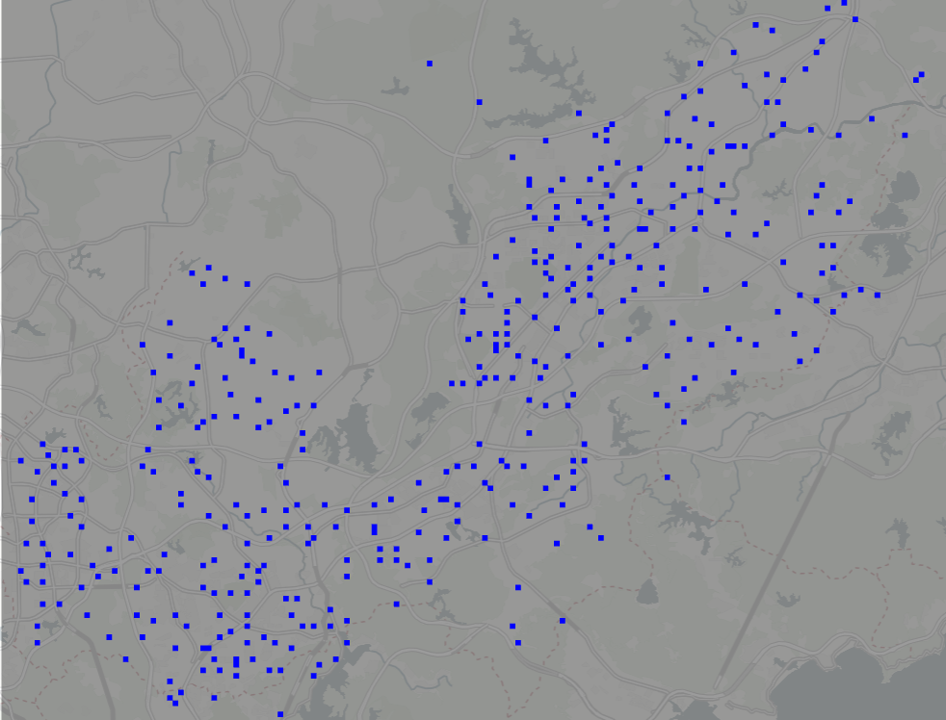}  
  \caption{Initial UAV Vertiport Configuration via $ k $-Means Clustering ($ k=400 $)}  
  \Description{A flat map visualizing 400 clustered UAV vertiport locations as translucent color blocks.}  
  \label{fig4-4}  
\end{figure}

Based on the population density of study area and Chinese low-altitude economy development plan, we estimated an initial requirement of 400 UAV vertiports. To generate a baseline configuration, we applied $ k $-means clustering ($ k=400 $) to UAV flight origin-destination pairs, using spatial coordinates as features. The resulting cluster centers, after gridding at 200-meter resolution, formed the initial planning map (Figure~\ref{fig4-4}), visualized as translucent color blocks representing potential vertiport locations.  

\begin{figure}[h]  
  \centering  
  \includegraphics[width=0.95\linewidth]{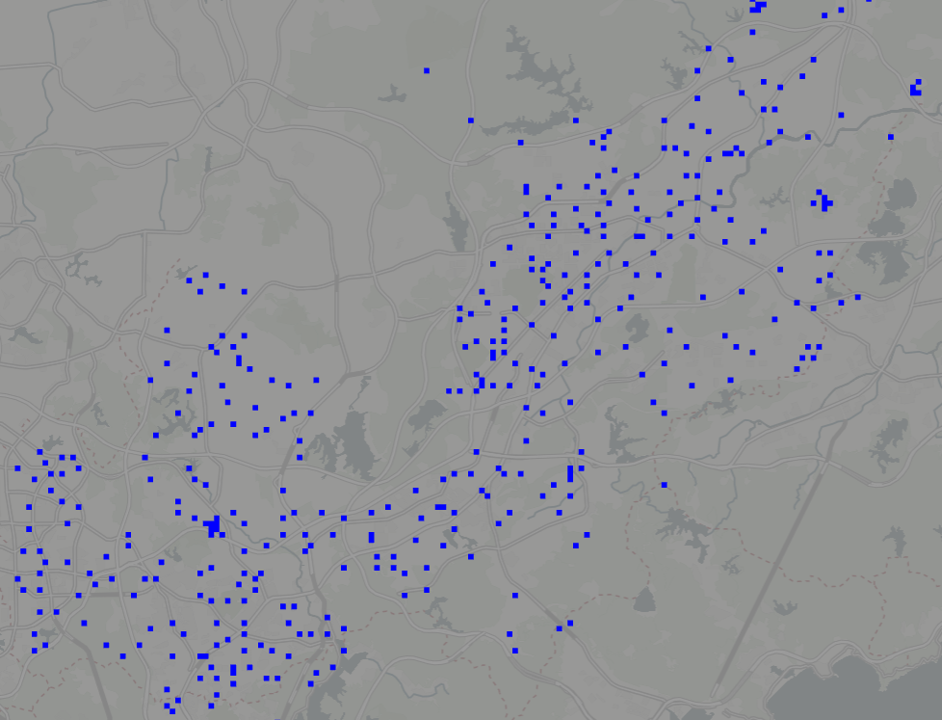}  
  \caption{Optimized UAV Vertiport Configuration After 300 Iterations}  
  \Description{A flat map showing optimized vertiport distribution with reduced spatial clustering compared to Figure~\ref{fig4-4}.}  
  \label{fig4-5}  
\end{figure}

The optimization model (Section~\ref{sec3}) was configured with a spatial resolution of 200 meters, temporal resolution of 1 hour, UAV vertiport service radius of 1 kilometer, and per-vertiport capacity capped at 20 service units per hour. After 300 optimization iterations—where one existing vertiport was relocated per round to balance demand fulfillment—the total unmet demand ($ \sum FRD $) decreased by 38\% (from 5421 to 3380). High-demand regions exhibited reduced vertiport clustering, while low-demand areas saw decommissioning of inefficient sites (Figure~\ref{fig4-5}), reflecting the algorithm’s prioritization of spatial efficiency.  

\begin{figure}[h]  
  \centering  
  \includegraphics[width=\linewidth]{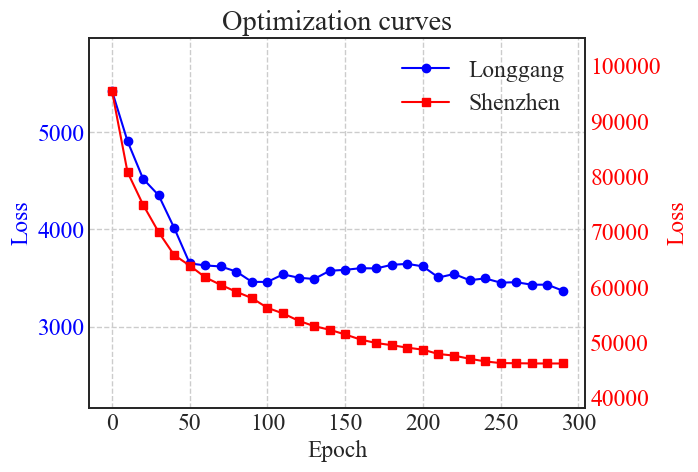}  
  \caption{Optimization Curves for the the study area (Blue) and the entire city (Red)}  
  \Description{A coordinate system plotting loss reduction over 300 iterations, with the red curve (entire city) showing steeper decline than the blue curve (study area).}  
  \label{fig4-6}  
\end{figure}

The optimization curve (Figure~\ref{fig4-6}) reveals diminishing returns beyond 300 iterations, with the loss value stabilizing as spatial adjustments saturate. Notably, when scaling to the entire city full 2000-vertiport plan, the loss decreased by 52\% (from 95445 to 46105), a more pronounced improvement attributed to two factors. First, larger-scale deployment smooths stochastic fluctuations, enabling more stable convergence. Second, the more complex terrain—particularly its mountainous regions (30\% of the area)—initially degraded performance due to sparse demand but allowed greater optimization gains through strategic reallocation.  

Three clustering algorithms—$ k $-means, Gaussian Mixture Model (GMM), and Hierarchical Agglomerative Clustering (HAC)—were tested with three initialization strategies. The first strategy directly clustered 400 vertiports, the second clustered 420 and pruned the smallest 20 clusters to filter sporadic demand, and the third clustered 500 and removed 100 clusters. Results showed that all methods achieved similar final losses ($ \pm 2\% $), suggesting that time-unaware clustering suffices for initial placement. However, a strong correlation ($ R^2 = 0.91 $) between initial and final loss values indicated the optimizer’s tendency to converge to local optima. Pruning-based strategies lifted losses by 22–35\% by inadvertently discarding valid demand signals when removing small clusters, underscoring the critical role of initialization quality in shaping algorithmic outcomes. A deeper limitation emerges: while clustering enables rapid initialization, the optimizer’s local convergence behavior inherently restricts its ability to escape suboptimal configurations. This suggests that future improvements, such as integrating reinforcement learning or metaheuristics, could focus on balancing computational efficiency with broader exploration of the solution space to achieve globally optimal vertiport networks.

\begin{table}[h]
  \caption{Experimental results of different initialization methods}
  \label{tab:tab1}
  \begin{tabular}{ccc} 
    \toprule
    Initialization method & Initialization loss & Optimized loss\\ 
    \midrule
    K-means 1 & 5421 & 3380\\ 
    K-means 2 & 5903 & 3793\\ 
    K-means 3 & 6646 & 4151\\ 
    GMM 1 & 5384 & 3348\\ 
    GMM 2 & 5914 & 3619\\ 
    GMM 3 & 6449 & 4060\\ 
    HAC 1 & 5603 & 3227\\ 
    HAC 2 & 6026 & 3824\\ 
    HAC 3 & 6527 & 4369\\ 
    \bottomrule
  \end{tabular}
\end{table}

\subsection{Results of recommendation system}

The recommendation system integrates diverse data sources, including subway network maps, UAV vertiport demand/supply distributions, population density, housing rental prices, and obstacle density. Demand for UAV vertiports is derived from Figure~\ref{fig4-5}, where a well-performing planning map (Section~\ref{sec3-3}) is gridded into a static demand tensor $ D \in \mathbb{Z}^{1 \times M \times N} $. Supply initialization leverages existing vertiport locations and dynamically incorporates user-planned additions. Population density and housing prices are sourced from street-level official statistics, while obstacle density is computed from altitude gradient maps.  

\begin{figure}[h]  
  \centering  
  \includegraphics[width=\linewidth]{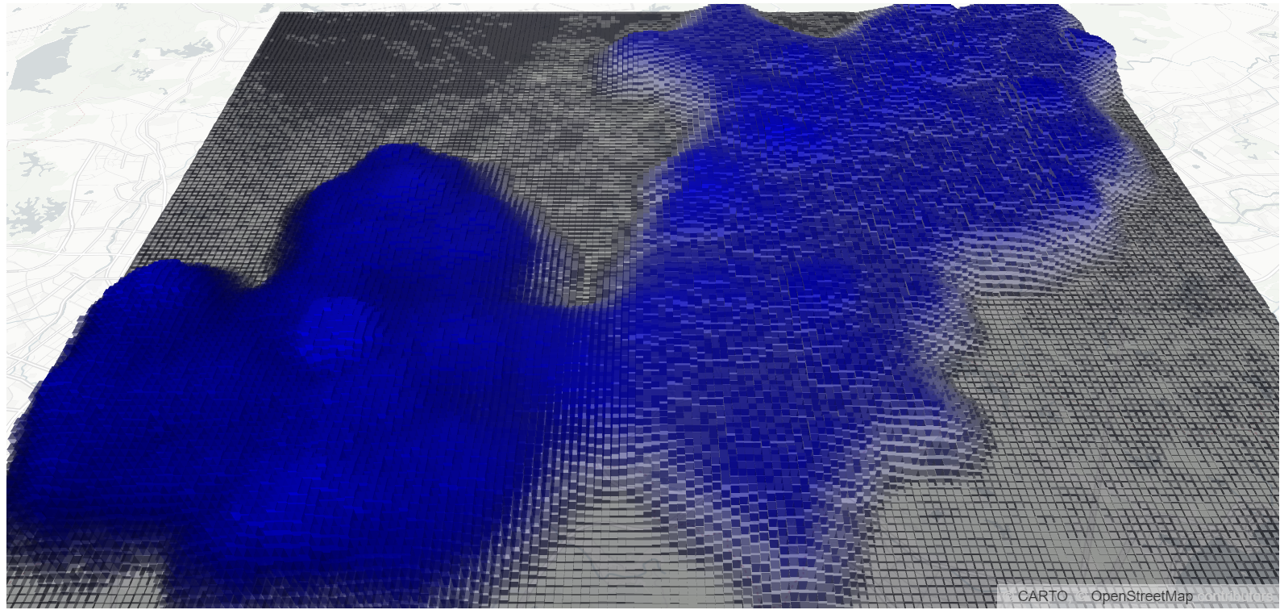}  
  \caption{Comprehensive Score Distribution at Initial Planning Stage}  
  \Description{A 3D map showing gridded regions with color-coded and height-scaled blocks representing initial comprehensive scores.}  
  \label{fig4-7}  
\end{figure}

The custom machine learning model synthesizes these inputs through four weighted scoring matrices. These include Score1 (maximizing demand satisfaction), Score2 (maximizing coverage), Score3 (minimizing air-ground connectivity time), and Score4 (minimizing construction costs, inversely weighted). After normalizing all scores, the model applies a weighted summation ($[1.0, 1.0, 1.0, -0.5]$) followed by a sigmoid activation to generate a comprehensive score matrix (Figure~\ref{fig4-7}). Early in the planning process, Score1 and Score4 dominate: high-demand regions (Figure~\ref{fig4-5}) align with elevated comprehensive scores, while construction costs (Score4) introduce localized fluctuations. This creates a demand-driven pattern with spatial undulations reflecting cost constraints.  

\begin{figure}[h]  
  \centering  
  \includegraphics[width=\linewidth]{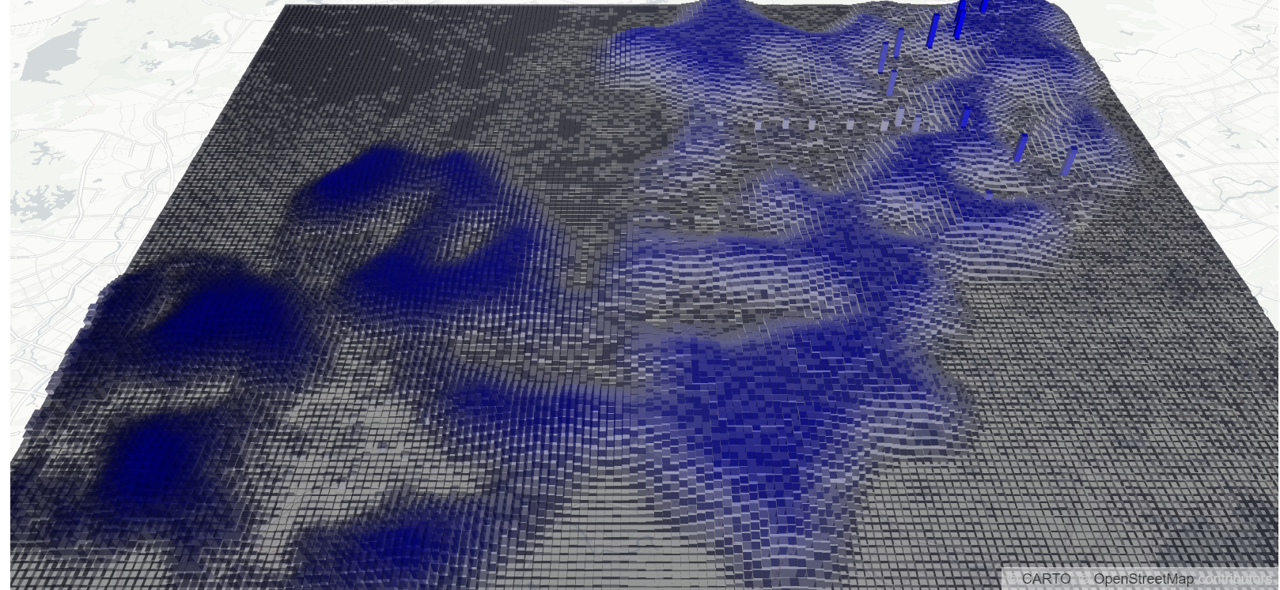}  
  \caption{Comprehensive Score Distribution in Later Planning Stages}  
  \Description{A 3D map visualizing score shifts toward peripheral regions as planning progresses.}  
  \label{fig4-8}  
\end{figure}

\begin{figure}[h]  
  \centering  
  \includegraphics[width=\linewidth]{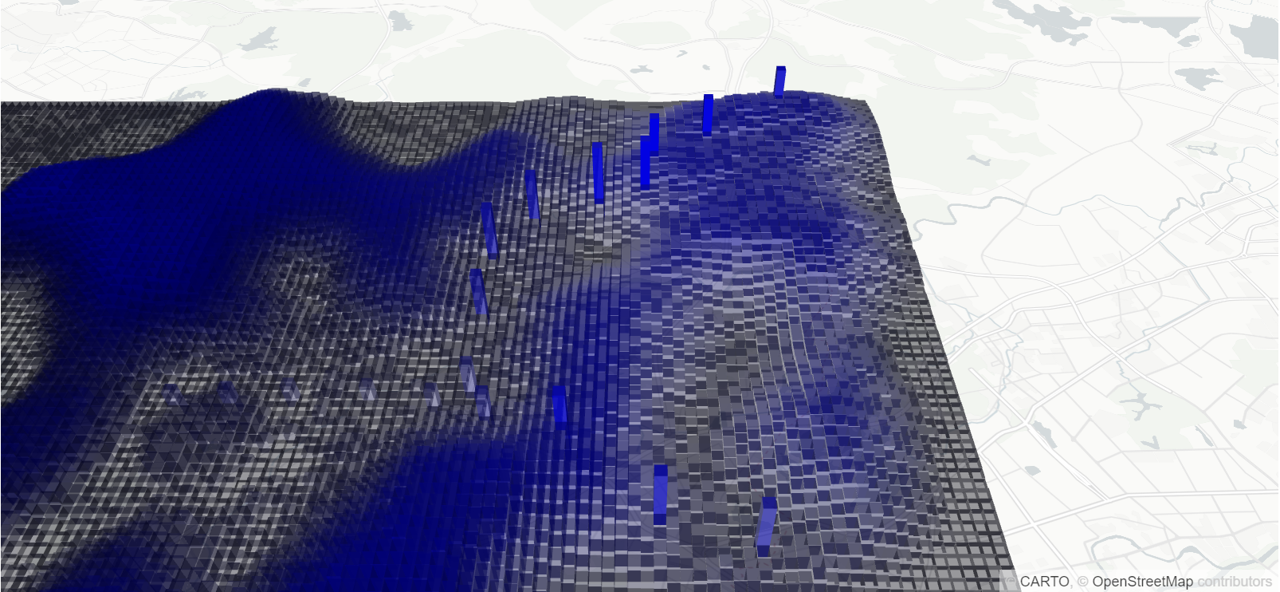}  
  \caption{High-Scoring Regions Correlated with Subway Stations}  
  \Description{A 3D map emphasizing clustered high scores near transit hubs.}  
  \label{fig4-9}  
\end{figure}

As planning progresses (Figure~\ref{fig4-8}), diminishing marginal returns on demand satisfaction reduce Score1’s influence, allowing Score2 to dominate. Peripheral areas with sparse demand but strategic coverage potential emerge as priority zones. Notably, Figure~\ref{fig4-9} highlights regions with persistently high scores corresponding to subway station locations. Constructing vertiports here significantly reduces air-ground connectivity time, an effect that intensifies as the network expands. This synergy becomes increasingly critical in later planning stages, where optimizing multimodal integration outweighs localized demand fulfillment.  

To validate the feedback mechanism’s effectiveness, we conducted an experiment where the custom machine learning model was initialized with parameters $[1.0, 1.0, 1.0, -0.5]$ (denoted as \textit{P1, P2, P3, P4}), but user selections prioritized recommendations generated using $[1.0, 0.5, 1.5, -1.0]$. Over 200 iterations, the model parameters adaptively evolved from initial values to $[0.91, 0.45, 1.23, -0.91]$. Crucially, the system’s performance hinges on the relative weights between parameters rather than their absolute magnitudes. When normalized, the final parameter ratios closely align with the user’s preference vector, demonstrating the feedback mechanism’s ability to learn and internalize decision-making patterns.  

\begin{figure}[h]  
  \centering  
  \includegraphics[width=\linewidth]{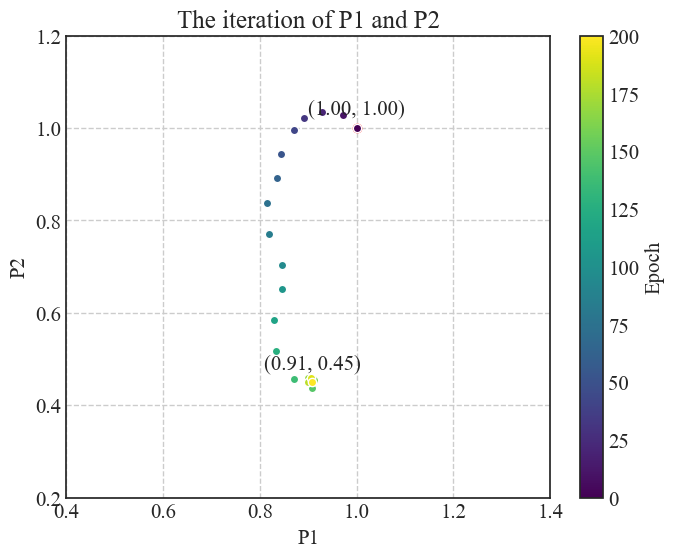}  
  \caption{Iterative Adjustment of Parameters \textit{P1, P2}.}  
  \Description{Side-by-side scatter plots showing parameter trajectories over 200 iterations, with color gradients indicating progression.}  
  \label{fig4-10}  
\end{figure}

\begin{figure}[h]  
  \centering  
  \includegraphics[width=\linewidth]{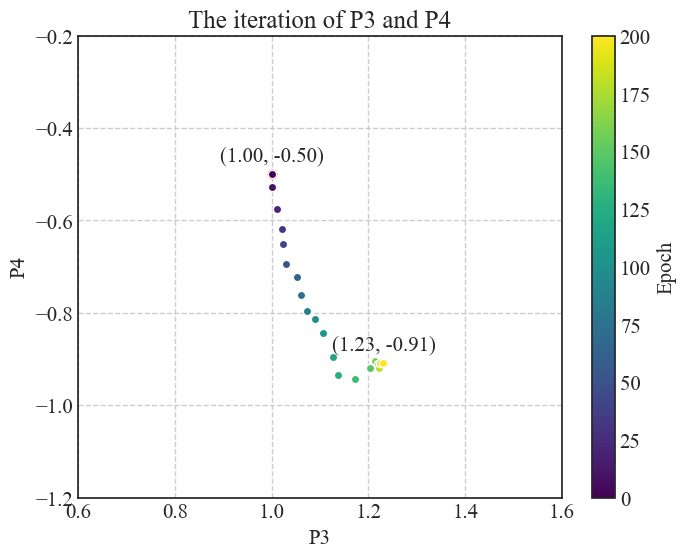}  
  \caption{Iterative Adjustment of Parameters \textit{P3, P4}.}  
  \Description{A coordinate system plotting $ P4 $ values over iterations, demonstrating asymptotic convergence to -0.92.}  
  \label{fig4-11}  
\end{figure}

Figures~\ref{fig4-10} and~\ref{fig4-11} visualize the iterative adjustment process. The first figure tracks the trajectory of \textit{P1, P2} over iterations, while the second illustrates \textit{P3, P4}’s convergence trend. Notably, \textit{P2} (coverage maximization weight) decreases most sharply, while \textit{P4} (construction cost penalty) nearly doubles in magnitude. This reflects the system’s adaptation to prioritize user behavior—favoring cost-efficient expansions and multimodal connectivity (as emphasized by the user’s preference for higher \textit{P3} and lower \textit{P2}). The results confirm that the feedback mechanism effectively bridges algorithmic outputs with human-in-the-loop decision-making, ensuring the recommendation system remains responsive to evolving planning priorities.  

\section{Discussion}
\label{sec5}

This study bridges theoretical modeling and practical deployment in urban air mobility (UAM) planning through three key innovations. First, while prior works often prioritize granular behavioral modeling (e.g., queuing dynamics, pricing mechanisms), they frequently simplify application scenarios and demand scales, limiting their applicability to large-scale urban environments. Conversely, classic covering location problems oversimplify user behavior by assuming binary service availability. Our CDMCLP synthesizes both perspectives: retaining the macro-level scalability of covering frameworks while incorporating micro-level user behavior through algorithm G and temporal demand dynamics. This dual focus enables the model to capture both systemic efficiency and localized decision-making, aligning with the low-altitude economy’s trajectory toward dense, heterogeneous UAV networks. Second, the recommendation system exemplifies the integration of academic rigor and operational practicality. By mathematizing ambiguous planning strategies, such as demand satisfaction, coverage maximization, and cost minimization, the system translates complex optimization results into interpretable scores for human planners. Key parameters (e.g., weight coefficients) remain user-adjustable, balancing algorithmic autonomy with domain expertise. While this introduces minimal cognitive load on planners, the trade-off proves negligible for high-stakes, large-scale infrastructure projects.

However, the proposed framework faces inherent challenges. Computational complexity escalates in megacity-scale deployments (e.g., selecting 2,000 vertiports from 140,000 candidates in Shenzhen), where traditional optimization algorithms struggle with combinatorial explosion. Additionally, the model’s reliance on high-resolution demand and urban context data raises data collection and preprocessing hurdles. For the recommendation system, challenges mirror those in personalized recommendation systems, such as adapting to shifting user priorities and balancing recommendation diversity with precisionm amplified by scarce user interaction data in UAM’s nascent stage.

\section{Conclusion}
\label{sec6}

This study advances large-scale UAM planning by integrating theoretical rigor with practical deployment considerations. Through the CDMCLP, we bridge macroscopic network evaluation with microscopic demand-supply interactions, enabling scalable performance assessment and optimization of UAV vertiport networks. The proposed framework’s dual focus, retaining the structural clarity of covering models while incorporating dynamic user behavior via algorithm G, positions it as a robust tool for high-density urban environments. Experimental validation in a Chinese center city demonstrates its efficacy: optimization algorithms combining clustering initialization and greedy updates reduced unmet demand by 38--52\%, producing high-performance planning maps that adapt to evolving socioeconomic and infrastructural constraints. Furthermore, the hybrid recommendation system bridges mathematical modeling and human decision-making, synthesizing competing objectives (demand coverage, cost efficiency, multimodal connectivity) into interpretable scores while retaining user control over critical parameters.

Despite these advancements, the framework’s full potential remains partially unrealized. The CDMCLP’s capacity to model heterogeneous demand types and UAV configurations, such as distinguishing commercial freight from passenger services, awaits empirical validation. Similarly, the optimization algorithm’s tendency toward local optima suggests opportunities for enhancement through metaheuristics or reinforcement learning, particularly in balancing computational efficiency with global exploration. For the recommendation system, future iterations could expand its scope by incorporating dynamic factors like meteorological conditions and policy incentives, while addressing data scarcity challenges in user preference modeling. These refinements will be critical as urban air mobility transitions from conceptual frameworks to operational realities, demanding tools that harmonize technical precision with adaptive, human-centric planning paradigms.





\bibliographystyle{ACM-Reference-Format}
\bibliography{main}


\end{document}